%% file: main.tex
\documentclass[10pt,twocolumn,letterpaper]{article}

\usepackage[pagenumbers]{cvpr} 

\input{preamble}
\definecolor{cvprblue}{rgb}{0.21,0.49,0.74}
\usepackage[pagebackref,breaklinks,colorlinks,allcolors=cvprblue]{hyperref}

\usepackage{multirow}
\usepackage{pifont}
\usepackage{makecell}
\usepackage{xcolor}
\usepackage{xspace}
\usepackage{algorithm}
\usepackage{algorithmic}
\usepackage{caption}
\usepackage{stfloats}
\definecolor{darkgreen}{rgb}{0.17,0.56,0.36}
\newcommand{\proposed}{UniLS\xspace}
\newcommand{\correct}{\textcolor{darkgreen}{\ding{51}}}
\newcommand{\wrong}{\textcolor{red}{\ding{55}}}

\definecolor{gold}{HTML}{FAE37F}
\definecolor{silver}{HTML}{D7D7D7}
\definecolor{bronze}{HTML}{EDBA91}
\newcommand{\au}[1]{\setlength{\fboxsep}{1pt}\colorbox{gold}{#1}}
\newcommand{\ag}[1]{\setlength{\fboxsep}{1pt}\colorbox{silver}{#1}}

\newcommand{\tablestyle}[2]{\setlength{\tabcolsep}{#1}\renewcommand{\arraystretch}{#2}\centering}

\def\paperID{****} 
\def\confName{CVPR}
\def\confYear{2026}

\title{\proposed: End-to-End Audio-Driven Avatars for Unified Listening and Speaking}

\author{%
Xuangeng Chu\textsuperscript{1,2}\thanks{Equal contribution.}\quad
Ruicong Liu\textsuperscript{1,2}\footnotemark[1] \thanks{Corresponding author.}\quad
Yifei Huang\textsuperscript{1,2} \quad
Yun Liu\textsuperscript{1} \quad
Yichen Peng\textsuperscript{3} \quad
Bo Zheng\textsuperscript{1} \\
\textsuperscript{1}Shanda AI Research Tokyo \quad
\textsuperscript{2}The University of Tokyo \quad
\textsuperscript{3}Institute of Science Tokyo \\
\texttt{\small \{xuangeng.chu, ruicong.liu, yifei.huang, yun.liu, bo.zheng\}@shanda.com} \\
\texttt{\small peng.y.ag@m.titech.ac.jp}
}

\begin{document}

\maketitle


\input{sec/0_abstract}
\input{sec/1_intro}
\input{sec/2_related}

\input{sec/3_motivation}
\input{sec/4_methods_experiments}

{
    \small
    \bibliographystyle{ieeenat_fullname}
    \bibliography{main}
}
\input{sec/X_suppl}

\end{document}

%% file: sec/0_abstract.tex
\begin{abstract}
Generating lifelike conversational avatars requires modeling not just isolated speakers, but the dynamic, reciprocal interaction of speaking and listening.
However, modeling the listener is exceptionally challenging: direct audio-driven training fails, producing stiff, static listening motions. This failure stems from a fundamental imbalance: the speaker's motion is strongly driven by speech audio, while the listener's motion primarily follows an internal motion prior and is only loosely guided by external speech. This challenge has led most methods to focus on speak-only generation. The only prior attempt at joint generation relies on extra speaker's motion to produce the listener.
This design is not end-to-end, thereby hindering the real-time applicability.
To address this limitation, we present \proposed, the first end-to-end framework for generating unified speak-listen expressions, driven by only dual-track audio.
Our method introduces a novel two-stage training paradigm.
Stage 1 first learns the internal motion prior by training an audio-free autoregressive generator, capturing the spontaneous dynamics of natural facial motion. Stage 2 then introduces the dual-track audio, fine-tuning the generator to modulate the learned motion prior based on external speech cues.
Extensive evaluations show \proposed achieves state-of-the-art speaking accuracy. More importantly, it delivers up to 44.1\% improvement in listening metrics, generating significantly more diverse and natural listening expressions. This effectively mitigates the stiffness problem and provides a practical, high-fidelity audio-driven solution for interactive digital humans.


\end{abstract}

%% file: sec/1_intro.tex
\begin{figure*}[htbp!]
	\centering
	\includegraphics[width=\linewidth]{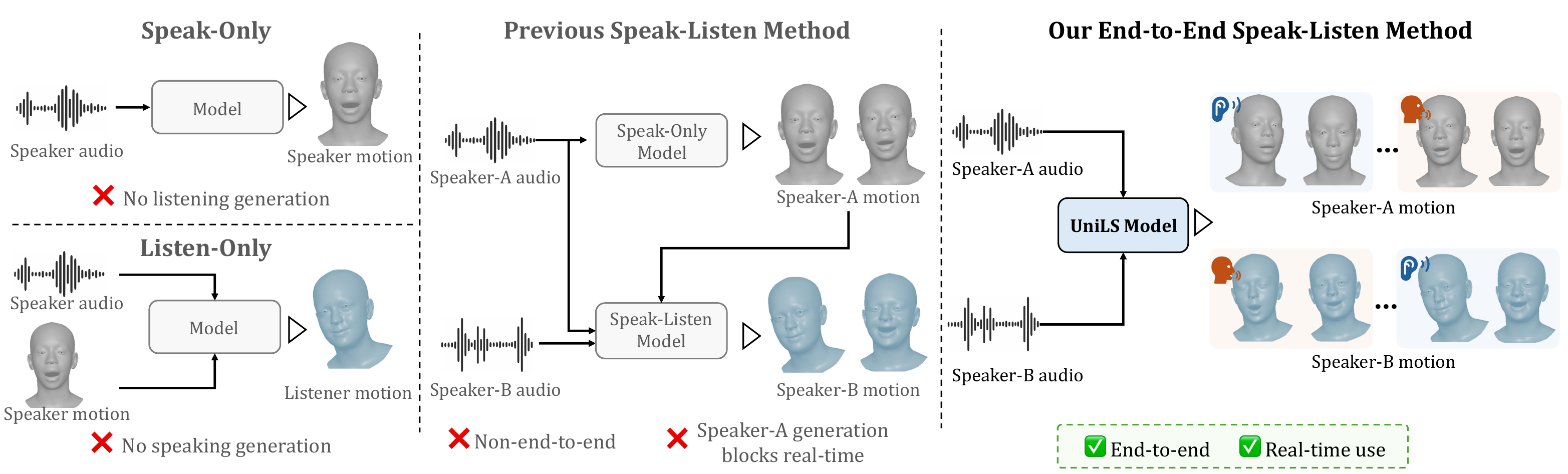}
	\caption{Comparison between previous methods and our proposed approach. Most previous studies remain one-way, \ie, speak-only or listen-only. The previous speak–listen method \cite{peng2025dualtalk} requires generating speaker A’s facial sequence before producing speaker B’s motions. The speaker-A generation makes it non-end-to-end and blocks real-time. In contrast, our method provides an end-to-end framework for unified, real-time speak–listen motion generation.}
	\label{fig:teaser}
\end{figure*}

\section{Introduction}
\label{sec:intro}
Conversational avatars \cite{niswar2009real,li2023one,peng2024synctalk,sung2024laughtalk,zhou2024meta} serve as a bridge between humans and intelligent virtual agents \cite{cassell2000human,diederich2022design,liao2023proactive,wu2024vgg}, enabling more natural and engaging human–computer interaction \cite{bai2023high,pang2023dpe,yu2023nofa,liu2024single,liu2025sfhand,gpavatar2024,chu2024gagavatar,lu2023audiodriven3dfacialanimation}.
Although recent studies have advanced this field, most systems remain one-way, \ie, either speak-only \cite{peng2023selftalk,peng2023emotalk,thambiraja2023imitator,codetalker2023} or listen-only \cite{liu2024customlistener,ng2023can,song2023react2023,tran2024dim}.
In real-world interaction, a conversational avatar should not only speak naturally but also listen naturally.
When the user is talking, it should show appropriate listening reactions (blinks, nods, subtle expressions), and when it speaks, its mouth and facial expressions should match its own voice.

Recently, the first attempt at building a conversational avatar \cite{peng2025dualtalk} has emerged, enabling joint speaking and listening facial motion generation.
As shown in \cref{fig:teaser}, it relies on speaker-A’s facial sequence to produce speaker-B’s speak–listen motions.
This dependency prevents the system from being fully end-to-end and requires an additional motion acquisition or generation pipeline.
The reliance on external motion input introduces extra processing stages and potential temporal lag, blocking the real-time deployment.

Unlike previous methods, we propose \proposed, an end-to-end framework for unified speaking and listening.
As illustrated in \cref{fig:teaser}, \proposed takes only audio inputs and generates the corresponding motions for both speaker-A and speaker-B.
A straightforward strategy is to jointly train an end-to-end model on “speak-listen” data.
However, this approach leads to a noticeable issue: the generated listening expressions become stiff and unnatural.
We empirically find that this arises from an imbalance in audio–motion correlation.
The speaker-A's speech strongly drives its own mouth and facial motion through phoneme and rhythm alignment, but only weakly relates to the speaker-B’s face.
Consequently, direct joint optimization drives the model to collapse into a safe, static facial prior for the listening branch, yielding expression stiffness.

To address this challenge, our key insight is to reinterpret listening behavior as a combination of internal motion prior and external audio cues.
Human listening expressions first follow their own motion patterns a priori (\eg, blink frequency, subtle nods, and muscle coordination), and then to be modulated by external speech.
Following this intuition, we introduce a two-stage training paradigm.
In stage 1, we remove audio input and train an audio-free generator on unpaired multi-scenario data.
The generator learns to predict the next facial motion, allowing it to build an internal motion prior.
In stage 2, we fine-tune the generator on paired conversational data.
The model is conditioned on speaker A and B’s speech, allowing the audio to drive facial motion.

To enable this research, we track FLAME \cite{FLAME2017} parameters for videos from the Seamless Interaction dataset \cite{si2025}, providing high-quality 3D facial motion supervision for both speaking and listening behaviors.
Evaluations demonstrate that \proposed achieves state-of-the-art accuracy in generating speaking motions.
More importantly, our method delivers substantial gains in listening quality, achieving up to 44.1\% improvement in distributional listening metrics over prior methods. This highlights the expressiveness and naturalness of our generated listening behaviors, resulting in more realistic and authentic conversational avatars.

Our contributions are as follows:

\begin{itemize}
\item We present the first end-to-end framework that enables a conversational avatar to both speak and listen naturally, driven solely by audio inputs.
\item We propose a two-stage training paradigm in response to the issue of listening stiffness.
\item Our method achieves state-of-the-art results, generating both high-quality speaking expressions and diverse, natural listening expressions.
\end{itemize}

%% file: sec/2_related.tex
\section{Related work}\label{sec:related}
\subsection{Speech-driven 3D facial animation}
Research on speech-driven 3D facial animation has evolved substantially over the past decades. Early works \cite{edwards2016jali,taylor2012dynamic,xu2013practical} relied on procedural or rule-based approaches, which segmented speech into phonemes and mapped them to visemes using handcrafted rules or constraint-based motion graphs. While these methods provided explicit control, they required laborious parameter tuning, mocap data specific to each subject, and offline processing, making them inflexible and often unable to capture the diversity and nuance of natural speech.

With the rise of deep learning, learning-based methods have largely replaced procedural pipelines by directly learning the mapping from audio to 3D facial motion. End-to-end architectures such as FaceFormer \cite{faceformer2022}, CodeTalker\cite{codetalker2023}, and SelfTalk \cite{peng2023selftalk} leverage mesh-based or vertex-level representations to generate expressive and temporally coherent animations. More recent frameworks, including UniTalker \cite{fan2024unitalker}, ScanTalk \cite{scantalk2024}, and ARTalk \cite{artalk2025} further improve cross-identity generalization and enable retargetable animation across diverse facial topologies. 
Despite these advances, most existing models focus solely on speaking behavior, ignoring speak-listen interactions that occur in natural conversations.

\subsection{Speak-listen 3D facial animation}
Recent research starts focusing on modeling nonverbal listener behaviors, which are essential for natural and engaging interactions.
DualTalk \cite{peng2025dualtalk} represents the first attempt to generate speak–listen facial motions within a unified framework, producing both speaking and listening behaviors. However, it depends on the interlocutor’s precomputed facial sequence, preventing true end-to-end generation and limiting real-time performance.
Earlier studies such as Learning2Listen \cite{ng2022learning} generated only listener reactions (\eg, nods or smiles) based on the speaker’s speech and facial mesh. While effective in producing feedback, these methods are limited to single-turn responses and do not support speak-listen interactions. Other methods \cite{ng2023can,wu2024vgg,zhou2022responsive} predicted listener facial expressions given conversational context, but their outputs remained confined to brief, isolated reactions without modeling temporal coherence.

\subsection{Autoregressive 3D facial animation}
Autoregressive approaches model facial motion as a temporal sequence, predicting each frame step-by-step conditioned on past motion and audio context. These methods often utilize pre-trained speech representation models such as wav2vec \cite{baevski2020wav2vec}, EnCodec \cite{defossez2024moshi}, or HuBERT \cite{hsu2021hubert} to extract expressive audio features, which are then mapped to 3D mesh vertices or facial model parameters. FaceFormer \cite{faceformer2022} autoregressively predicts continuous facial motion parameters to ensure smooth temporal coherence, while MeshTalk \cite{richard2021meshtalk} models discrete expression tokens to capture diverse articulation patterns. CodeTalker \cite{codetalker2023} and MultiTalk \cite{sung2024multitalk} adopt a VQVAE-based \cite{van2017neural} tokenization of facial motion and employ transformer decoders to predict future motion tokens from audio cues. Learn2Talk \cite{zhuang2024learn2talk} enhances FaceFormer with auxiliary lip-reading and lip-sync losses for improved alignment between phonemes and lip movements. MMHead \cite{wu2024mmhead} introduces textual conditioning for semantically guided motion control. More recently, ARTalk \cite{artalk2025} advances this field by combining autoregressive modeling with style priors, achieving high-fidelity and temporally stable 3D facial animation from speech.

%% file: sec/3_motivation.tex
\begin{figure*}[htbp!]
	\centering
	\includegraphics[width=\linewidth]{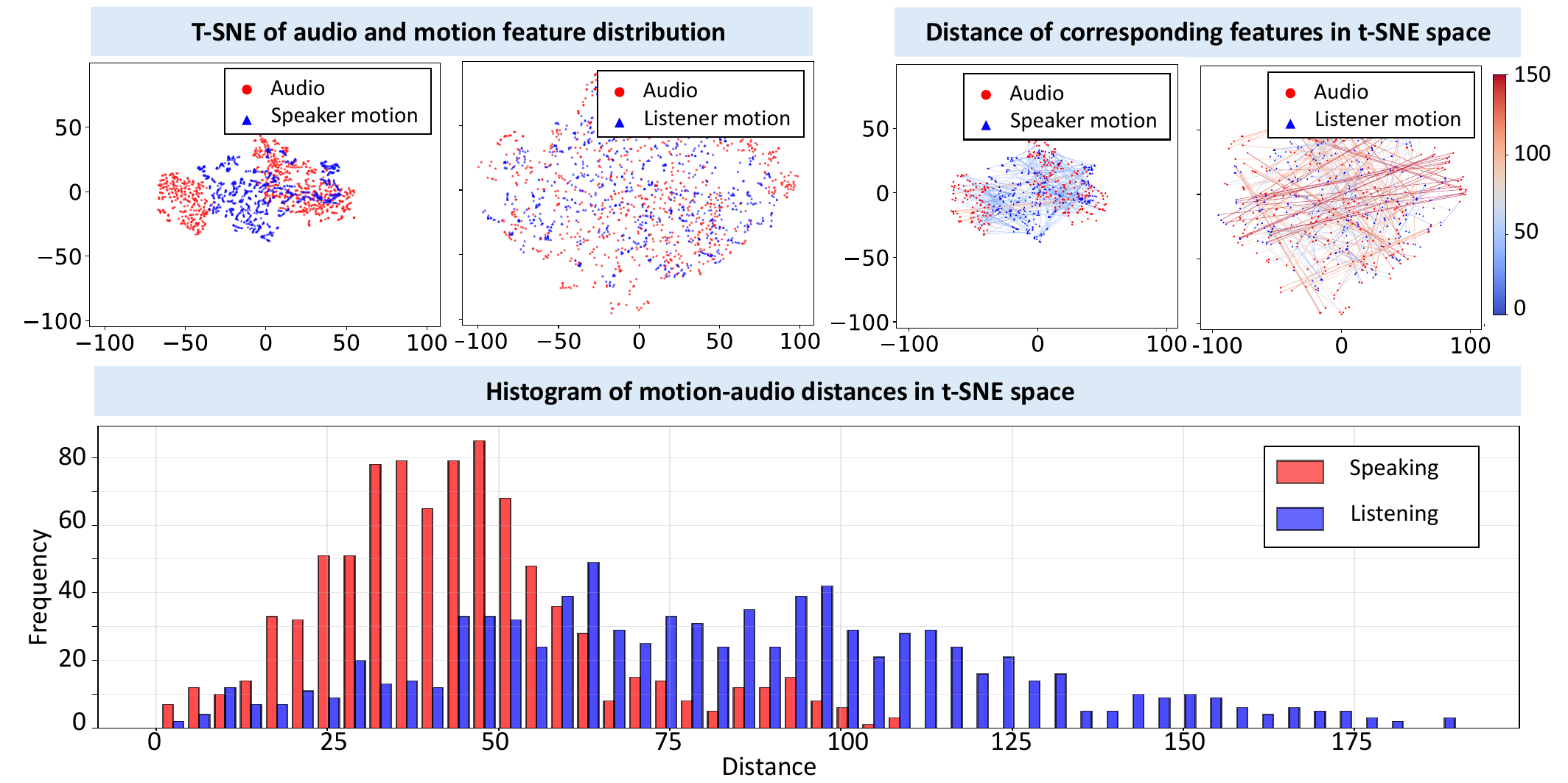}
	\caption{Correlation between facial expression parameters \cite{FLAME2017} and corresponding audio features \cite{baevski2020wav2vec}. For speaking, the audio is the speaker’s own speech. For listening, the audio comes from the other speaker’s speech.}
	\label{fig:correlation}
\end{figure*}

\section{Motivation}

\textbf{Observation: stiffness in listening motion.} When training a conversational avatar in an end-to-end manner, we observe a clear issue: the listening expressions become stiff whenever their own audio is silent.
For example, when speaker-A is listening while speaker-B is talking, speaker-A’s audio contains only silence.
The model learns to generate minimal or no motion, causing speaker-A’s listening behavior to degrade into a poker-face expression with little blinking, negligible micro-expressions, and almost no head movement.
In contrast, a natural conversational avatar should exhibit active listening cues, such as subtle nods, timely blinks, shifts in eye gaze, and soft facial movements that reflect attention and engagement.

\noindent\textbf{Analysis: low correlation between audio and listening motion.} To understand why naive end-to-end training produces stiff listening behaviors, we analyze the correlation between audio features \cite{baevski2020wav2vec} and expression parameters \cite{FLAME2017} for both speaking and listening.
As shown in \cref{fig:correlation}, the t-SNE distributions reveal a clear pattern: audio features cluster closely with speaking motions but show much weaker alignment with listening motions.
In the speaking case, audio and motion embeddings occupy nearby regions, with consistently small pairwise distances.
In contrast, listening motions are scattered farther from the corresponding audio features. This indicates that listening behavior is only weakly correlated with the audio of the interlocutor, as many aspects of natural listening, such as blinking or micro-expressions, occur independently of the speech signal.
This imbalance explains the listening stiffness: the network easily learns strong audio-driven mappings for speaking but receives insufficient guidance for listening, causing the listening branch to collapse into safe, low-variance motions.

\noindent\textbf{Insight.} According to the analysis, listening behavior should not be learned as a direct audio-to-motion mapping.
Instead, natural listening arises from two components:
1) an internal motion prior that reflects spontaneous behaviors such as blinks, nods, and micro-expressions,
and 2) external audio cues that modulate these intrinsic dynamics in response to conversational context.
We design a two-stage training framework that separately learns these two components.
In Stage 1, we train an audio-free generator to model the internal dynamics of facial behavior.
In Stage 2, we finetune this generator by conditioning on dual-track audios, allowing external speech signals to modulate the facial expression.

%% file: sec/4_methods_experiments.tex
\begin{figure*}[htbp!]
	\centering
	\includegraphics[width=\linewidth]{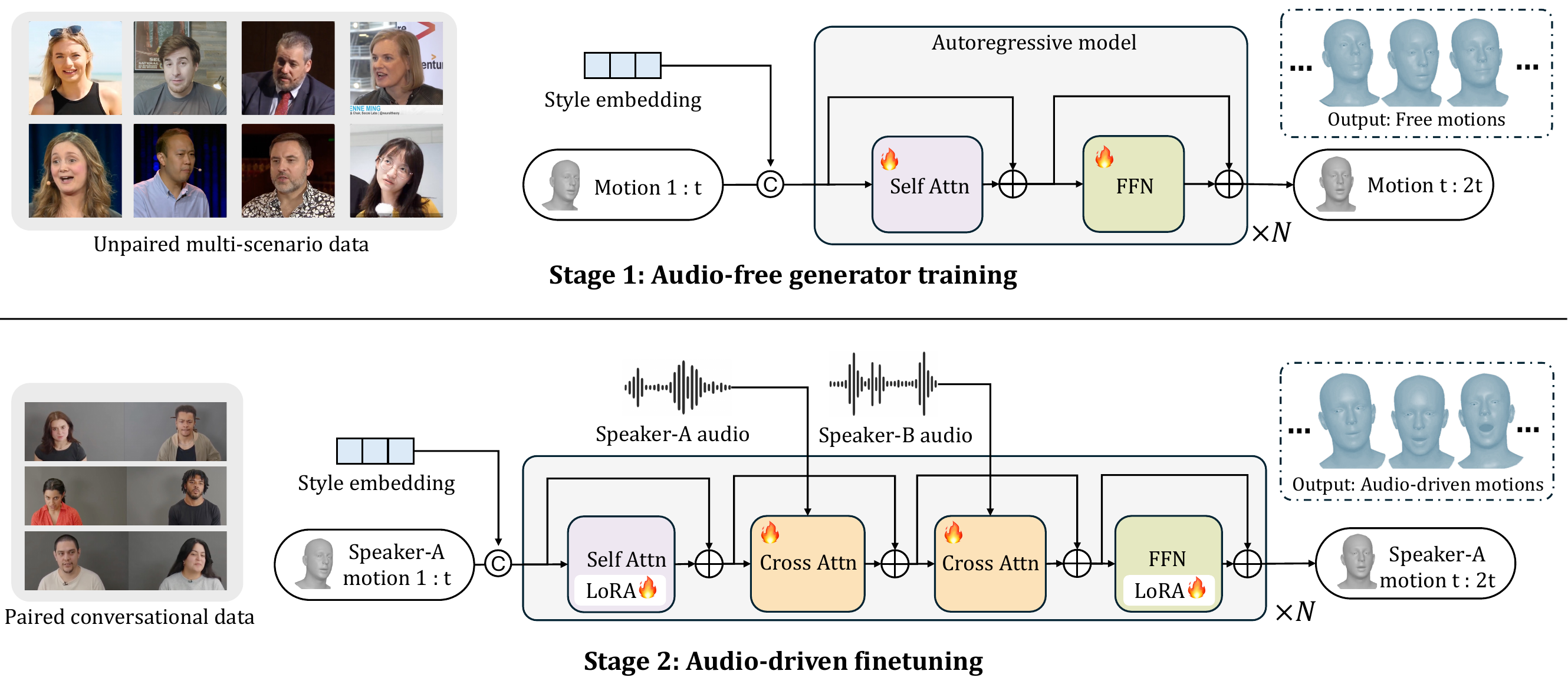}
	\caption{Overview of our two-stage training strategy. Stage 1 trains an audio-free generator on unpaired multi-scenario video data without using audio. Given past motions and a style embedding, the model predicts future free motion chunks. Stage 2 finetunes the generator on paired conversational clips by conditioning on speaker-A and speaker-B’s audios through cross-attention, producing audio-driven speak–listen motions.}
	\label{fig:method}
\end{figure*}

\section{Methods}
\label{sec:methods}
An overview of our method is presented in \cref{fig:method}.
We train a chunk-based autoregressive model to generate facial motions, conditioned on both speakers' audio.
The following sections provide further details: \cref{sec:41} describes the problem definitions, \cref{sec:42} explains the multi-scale codec model, \cref{sec:43,sec:44} introduce the two-stage training of our speak-listen model.

\subsection{Preliminaries}
\label{sec:41}
We represent facial motion using the FLAME 3D deformable model \citep{FLAME2017}, a widely adopted parametric head model. FLAME encodes dynamic facial behavior through expression parameters \(\psi\), poses parameters \(\theta\) (including head pose, jaw pose and eye gaze \cite{liu2024pnp,huang2020mutual}), and shape parameters \(\beta\).
Given these parameters, FLAME reconstructs a full 3D face mesh with 5,023 vertices $V$ using standard blendshape deformation and articulated pose transformations.
For each motion chunk, we form a motion matrix $M\in\mathbb{R}^{T\times D}$ by concatenating the expression $\psi$ and pose $\theta$ over $T$ consecutive frames.
Our goal is to learn an autoregressive generator $\mathcal{G}$ that takes audios as input and produces unified speak–listen facial motions coherently and naturally.

\subsection{Multi-scale codec} \label{sec:42}
Following previous work \cite{artalk2025}, we adopt a temporal multi-scale VQ codec to obtain a compact yet expressive discrete representation of 3D facial motion. This codec has been proven to achieve high-fidelity synthesis
with strong temporal coherence.

Given an input motion chunk $M$, a transformer encoder first extracts a temporal feature $\mathbf{f}$ into a latent space. This latent feature is then quantized using a multi-scale codebook.
Specifically, the codebook has a sequence of scales $[k_1, k_2, ..., k_L]$, where each level progressively refines the temporal resolution.
At the $l$-th scale, the feature $\mathbf{f}^{(l)}$ are quantized and interpolated to motion code $\mathbf{c}^{(l+1)}$, following:

\begin{equation}
\begin{aligned}
\mathbf{c}^{(l+1)} &= \text{Interp}(\text{Quant}(\mathbf{f}^{(l)}), k_l),\\
\mathbf{f}^{(l+1)} &= \mathbf{f}^{(l)} - \mathbf{c}^{(l+1)},
\end{aligned}
\label{eq:codec}
\end{equation}

\noindent where feature $\mathbf{f}^{(0)}$ is from the original encoder output.
Function $\text{Quant}(\cdot)$ maps features to the nearest codebook entries, while $\text{Interp}(\cdot, k_l)$ adjusts temporal resolutions to ensure smooth transitions.
We accumulate motion code $\mathbf{c}^{(l)}$ across scales to represent the original motion.
These multi-scale discrete motion codes serve as the supervision targets for training our autoregressive generator.
For clarity in the following subsections, we directly describe the model using the continuous motion matrix $M$, while the underlying codec operates in the discrete code space.

\subsection{Stage 1: Audio-free generator training}\label{sec:43}
In the first stage, we train an audio-free generator to learn internal motion priors that capture natural facial dynamics independent of speech.
This generator is trained on unpaired multi-scenario data, including diverse video sources such as news broadcasts, interviews, streaming content, and casual talking videos, providing diverse facial behaviors across identities and environments.

As shown in \cref{fig:method}, the input contains motion chunk $M$ with style embedding $\mathbf{s}$.
This style embedding is learned to encode speaker-specific motion characteristics, following \cite{artalk2025}.
This embedding is concatenated with the input chunk and then fed them into our autoregressive model $\mathcal{G}$.
Our model is transformer-based, composed of stacked self-attention and feed-forward blocks.
At each time step $t$, the generator predicts the next-chunk motion based solely on past motions and the style embedding, \ie:

\begin{equation} \label{eq:free-gen}
\hat{M}_{t:2t}=\mathcal{G}(M_{1:t}, \mathbf{s}).
\end{equation}

\noindent The model is trained with an autoregressive reconstruction loss over each chunk:

\begin{equation} \label{eq:loss}
\mathcal{L}=\sum_{t=1}^T || \hat{M}_{t:2t} - M_{t:2t}||.
\end{equation}

\noindent Through this process, the model learns to produce free motions, which reflects the internal motion prior such as blinking, subtle head movements, and micro-expressions.

\subsection{Stage 2: Audio-driven finetuning} \label{sec:44}
In the second stage, we finetune the generator to produce audio-driven conversational motions, enabling both speaking and listening behaviors.
This stage uses paired conversational data, where synchronized videos and audios from speaker-A and speaker-B provide the appropriate dynamics required for natural dialogue modeling.

Here we describe the process for generating speaker-A’s motion as an example. As illustrated in \cref{fig:method}, the input consists of three components: the motion chunk $M$, style embedding $\mathbf{s}$, and the audios $\mathbf{a}^A, \mathbf{a}^B$ from speaker-A and speaker-B, respectively.
To incorporate audio guidance, we extend the stage 1 architecture by adding two cross-attention layers to each transformer block.
Specifically, one attends to speaker-A’s audio (for speaking behavior) and the other attends to speaker-B’s audio (for listening behavior).
The newly added cross-attention layers are trained from scratch, while the backbone weights inherited from stage 1 are finetuned with LoRA \cite{hu2022lora}.
This design ensures allows the model to adapt efficiently to audio conditioning without overwriting the learned internal motion priors.
Formally, the generating process of stage 2 is expressed as:

\begin{equation} \label{eq:audio-gen}
\hat{M}_{t:2t}=\mathcal{G}(M_{1:t}, \mathbf{a}^A_{1:t}, \mathbf{a}^B_{1:t}, \mathbf{s}).
\end{equation}

\noindent The training objective remains a chunk-wise autoregressive reconstruction loss, which is the same as \cref{eq:loss}.
Generating motions for speaker-B follows the same procedure, with the two audios simply exchanged in their roles.
Through this finetuning, the generator combines internal motion priors learned in stage 1 with dual-track audio guidance, producing smooth and expressive speak–listen motions.

\noindent\textbf{Implementation Details.}
Our codec model's codebook consists of 256 entries, each with a code dimension of 64.
The time window size is 100 frames (4 seconds), and the multi-scale levels are [1, 5, 25, 50, 100].
We used the AdamW optimizer with a learning rate of 1.0e-4 for training the codec, with a total batch size of 64 for 100,000 iterations.
In the two-stage training, we train the autoregressive model using the AdamW optimizer with the same learning rate of 1.0e-4 and a batch size of 128 for 200,000 iterations.
During this training, we employ a frozen wav2vec audio encoder \cite{baevski2020wav2vec}.
All training was conducted on four NVIDIA H200 GPUs, requiring a total of approximately 40 GPU hours (10 GPU hours for the stage 1 and 30 GPU hours for stage 2).

\section{Experiments}
\label{sec:experiments}

\begin{table*}
    \caption{
        Evaluation of speaking and listening facial motions on the Seamless Interaction \citep{si2025} test split.
        We use colors to denote the \au{first} and \ag{second} places, respectively.
    }
    \label{tab:main_res}
    \tablestyle{8.9pt}{1.05}
    \centering
    \small
    \begin{tabular}{l|ccccc|ccccc}
    \toprule[1.2pt]
    & \multicolumn{5}{c|}{Speak} & \multicolumn{5}{c}{Listen} \\
    Method      & LVE$\downarrow$ & MHD$\downarrow$  & FDD$\downarrow$  & PDD$\downarrow$ & JDD$\downarrow$ & FDD$\downarrow$  & PDD$\downarrow$ & JDD$\downarrow$ & F-FID$\downarrow$  & P-FID$\downarrow$  \\
    \midrule
    DiffPoseTalk~\citep{diffposetalk2024} & 9.48 & 2.96 & 32.66 & 7.89 & 1.40 & - & - & - & - & -\\
    ARTalk~\citep{artalk2025} & 7.46 & 2.12 & 31.64 & \ag{7.66} & 1.19 & - & - & - & - & - \\
    ARTalk*~\citep{artalk2025} & 6.79 & 2.02 & \ag{27.41} & 8.55 & \ag{0.81} & \ag{30.62} & \ag{9.52} & \ag{1.53} & \ag{10.779} & \ag{0.072}  \\
    DualTalk~\citep{peng2025dualtalk} & \ag{6.35} & \ag{1.95} & 37.46 & 9.70 & 1.02 & 43.58 & 10.71 & 2.02 & 13.143 & 0.079 \\
    \midrule
    \proposed (ours) & \au{5.83} & \au{1.89} & \au{18.41} & \au{4.67} & \au{0.71} & \au{17.12} & \au{4.75} & \au{0.98} & \au{4.304} & \au{0.038} \\
    \bottomrule[1.2pt]
    \end{tabular}
\end{table*}

In this section, we first introduce the details of the dataset, provide an overview of the implementation of our method, describe the metrics used, and present the baseline methods.
Subsequently, we compare our method with existing approaches across a range of evaluation metrics.

\subsection{Experiment setting}
\noindent\textbf{Datasets.} 
For paired conversational data, we use the Seamless Interaction dataset \citep{si2025}, which offers large-scale dyadic conversational videos.
For multi-scenario data, we additionally adopt four large-scale video datasets: CelebV, TalkingHead-1KH \cite{wang2021one}, TEDTalk \cite{chung2016lip}, and VFHQ \cite{xie2022vfhq}.
To enable 3D facial motion supervision, we apply a carefully designed tracking pipeline to extract per-frame FLAME \cite{FLAME2017} parameters, including detailed eye-gaze \cite{liu2024uvagaze,liu2025gaze} and head pose annotations.
After filtering, we obtain 657.5 hours of conversational data from Seamless Interaction, and 546.5 hours of multi-scenario data from other datasets. The conversational data includes 251.5 hours of speaking motions comprising 22.6M frames, and 406.0 hours of listening motions comprising 36.5M frames. 
For the conversational dataset, we use 622.5 hours for training, 4.8 hours for validation, and 30.2 hours for testing.



\noindent\textbf{Evaluation Metrics.}
Our metrics evaluate both speaking accuracy and listening naturalness.

\noindent \textit{a) Speaking.} To assess lip synchronization quality, we use Lip Vertex Error (LVE) \citep{richard2021meshtalk}, which measures the maximum per-frame lip-vertex error. To assess overall facial accuracy, we use Mean Head Distance (MHD), the average distance across all head vertices. To assess upper-face motion consistency, we use Upper-face Dynamic Deviation (FDD) \citep{codetalker2023}. To assess temporal pose dynamics, we use Pose Dynamic Deviation for head pose (PDD) and jaw pose (JDD).

\noindent \textit{b) Listening.} For listening behaviors, we compute FDD, PDD, and JDD on listening segments to assess whether the generated motion dynamics follow the ground-truth distribution. We additionally report FID on FLAME expression and pose parameters to assess distributional naturalness.

\subsection{Quantitative results}
To evaluate our method's performance, we compare \proposed with representative baselines:

\begin{itemize}
\item \textit{DiffPoseTalk} \cite{diffposetalk2024}: A diffusion-based transformer model to generate speaking expressions and poses.
\item \textit{ARTalk} \cite{artalk2025}: A autoregressive model to generate speaking expressions and poses.
\item \textit{ARTalk*}: We adapt the original ARTalk for speak-listen generation by adding additional audio input.
\item \textit{DualTalk} \cite{peng2025dualtalk}: A non-end-to-end speak-listen model.
\end{itemize}

\cref{tab:main_res} reports an evaluation of speaking and listening facial motions.
Our method shows clear improvements in lip-sync accuracy (LVE, MHD) and speech-style alignment (FDD, PDD, JDD).
These results indicate that our model not only tracks phoneme-to-motion correspondence precisely but also captures the characteristic dynamics of speech, such as upper-face involvement and coordinated head-jaw movement.
For listening, our approach shows large improvements in distributional measures (FDD, PDD, JDD, F-FID, P-FID). 
This indicates that our model generates diverse expressions and head movements instead of collapsing into a neutral or static listening pose.
Together, these results validate our two-stage design, showing that the model successfully learns to produce both natural listening reactions and accurate speaking expressions within a unified framework.

\begin{figure}
\begin{center}
\centerline{\includegraphics[width=.845\linewidth]{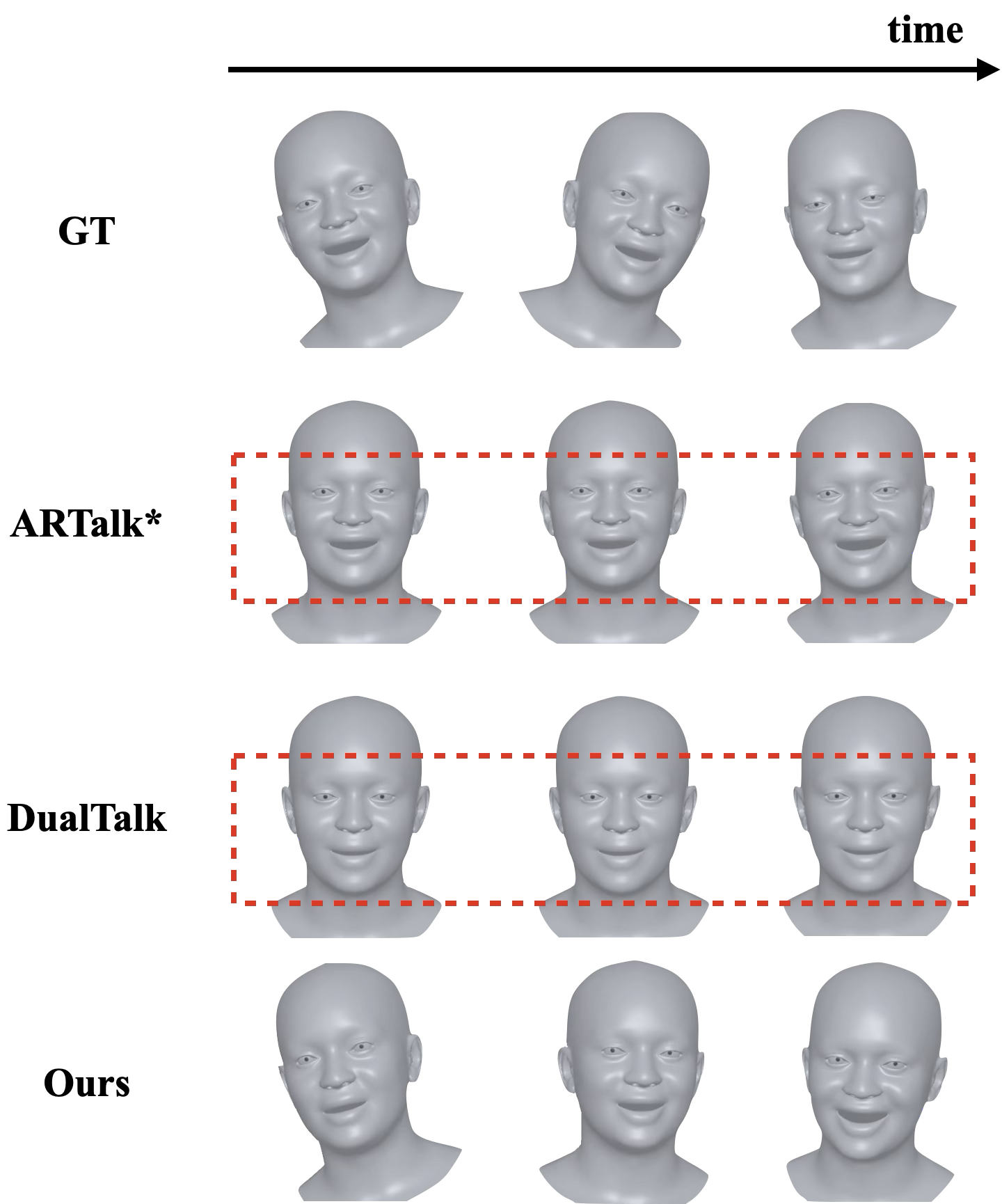}}
\caption{
Qualitative comparison on listening motions. \textcolor{red}{Red rectangles} highlight motion stiffness over time. Additional qualitative evaluation results are available in the supplementary materials.
}
\label{fig:main_results_listen}
\end{center}
\vspace{-1.5em}
\end{figure}

\begin{figure*}[ht]
\begin{center}
\centerline{\includegraphics[width=0.9\linewidth]{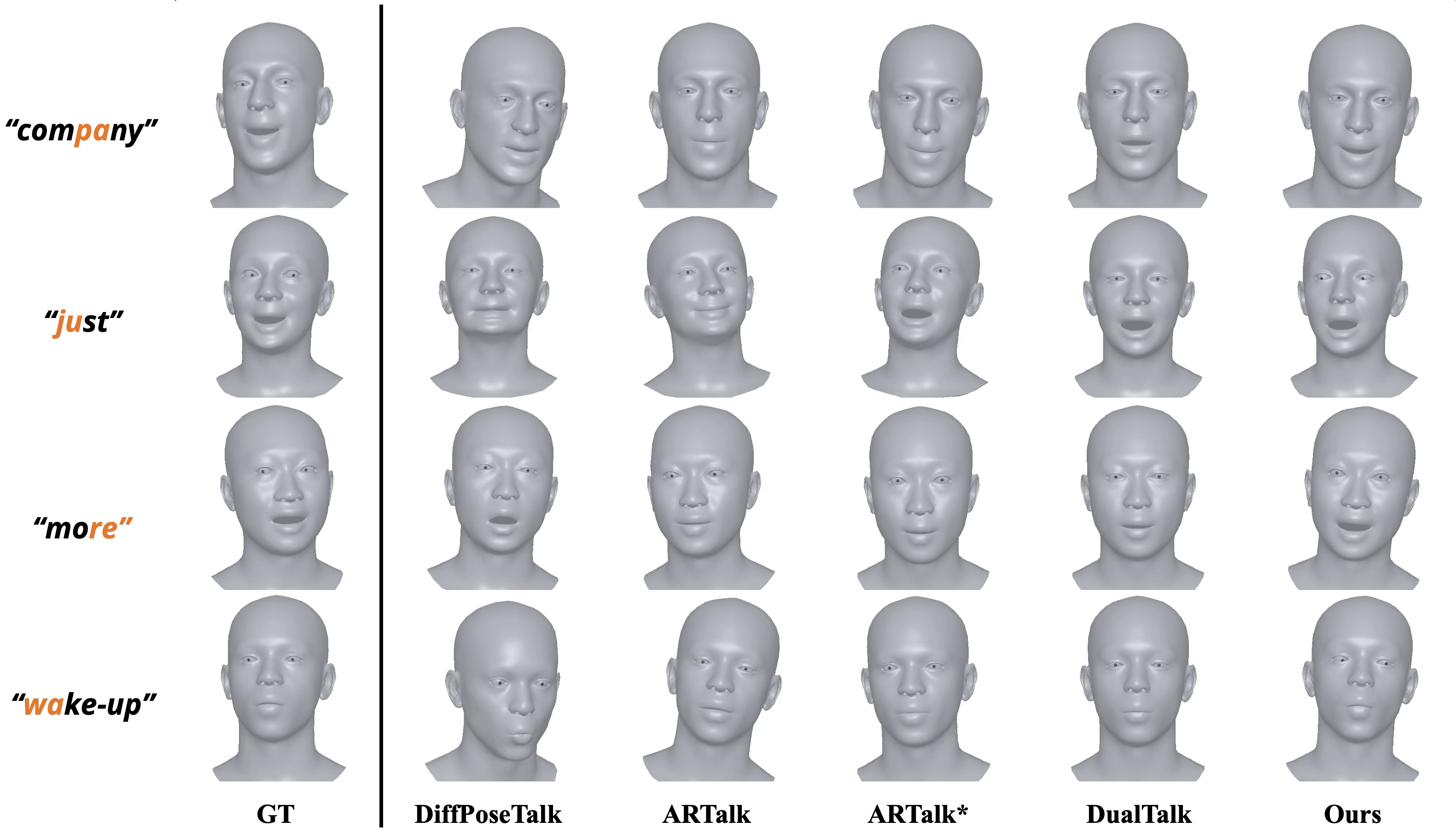}}
\caption{
Qualitative comparison on speaking motions.
Our method shows better alignment with the ground truth in expression style and lip synchronization. Additional qualitative evaluation results are available in the supplementary materials.
}
\label{fig:main_results_speak}
\end{center}
\end{figure*}

\subsection{Qualitative results}

In \cref{fig:main_results_speak,fig:main_results_listen}, we qualitatively compare our method with other baseline methods for listening and speaking motions, respectively.
As shown in \cref{fig:main_results_listen}, we show the ability of our method on listening motion generation.
ARTalk* and DualTalk exhibit noticeably stiff and low-variance facial motions, as highlighted by the red dashed boxes.
In contrast, our method produces vivid, expressive, and temporally diverse expressions. 
The generated faces exhibit natural head dynamics, mouth shapes, and micro-expressions, demonstrating the effectiveness of our two-stage training in capturing realistic listening behavior.
In \cref{fig:main_results_speak}, our method demonstrates excellent lip synchronization, accurately capturing various phonetic elements.
Furthermore, the generated results exhibit realistic facial behavior, such as micro-expressions and head movements.
Additional qualitative results are available in supplementary materials.

\begin{table}[t]
\caption{
User study results with 25 participants. Numbers (\%) indicate the proportion of users who prefer our method over each baseline. “Sync” measures lip synchronization, while “Exp”, “React”, and “Pose” assess the naturalness of facial expressions, listening reactions, and head pose, respectively.
}
\label{tab:user_study}
\tablestyle{7.5pt}{1.00}
\centering
\small
\begin{tabular}{lcccc}
\toprule[1.2pt]
Method & Sync  & Exp  & React  & Pose \\
\midrule
vs. DiffPoseTalk~\citep{diffposetalk2024}    & 55.34 & 61.65 & - & 58.25 \\
vs. ARTalk~\citep{artalk2025}                & 75.36 & 75.36 & - & 71.98 \\

vs. ARTalk*~\citep{artalk2025}           & 76.92 & 77.88 & 79.80 & 74.52 \\
vs. DualTalk~\citep{peng2025dualtalk}        & 86.06 & 90.38 & 91.35 & 89.42 \\
\bottomrule[1.2pt]
\end{tabular}
\end{table}
\subsection{User study}
User studies are recognized as an essential and reliable method for evaluating the performance of generative models.
To thoroughly assess our method, we conducted a user study examining four key aspects of conversational facial motion: lip synchronization, facial expression naturalness, listening reaction naturalness, and head pose naturalness. 
For baseline methods that do not generate listening behaviors, the reaction naturalness category is omitted.
We adopt a pairwise comparison protocol. For each trial, videos generated by our method and a baseline model are shown side by side in a randomized order. After viewing the video pairs, participants select the result they find more natural and realistic.
We then compute the percentage of users who prefer our method in each category.

As summarized in \cref{tab:user_study}, our method is consistently preferred over all baselines across all aspects.
Among the 25 participants in our study, the most notable improvement appears in listening reactions, where over 90\% of participants prefer our results compared to DualTalk.
This overwhelming preference highlights the strength of our two-stage design: our model produces listening motions that are significantly more expressive, responsive, and human-like than existing methods.

\begin{table*}
    \caption{
        Ablation study on different components in our framework. 
    }
    \label{tab:ablation}
    \tablestyle{8pt}{1.05}
    \centering
    \small
    \renewcommand\arraystretch{1.1}
    \begin{tabular}{l|ccccc|ccccc}
    \toprule[1.2pt]
    & \multicolumn{5}{c|}{Speak} & \multicolumn{5}{c}{Listen} \\
    Method      & LVE$\downarrow$ & MHD$\downarrow$  & FDD$\downarrow$  & PDD$\downarrow$ & JDD$\downarrow$ & FDD$\downarrow$  & PDD$\downarrow$ & JDD$\downarrow$ & F-FID$\downarrow$  & P-FID$\downarrow$  \\
    \midrule
    \textit{single cross-attn} & 11.48 & 3.00 & 31.35 & 5.83 & 0.92 & 27.46 & 7.01 & 1.61 & 5.97 & 0.059 \\
    \textit{w/o stage 1}  & 6.32 & 2.02 & 25.10 & 5.76 & 0.89 & 25.64 & 6.66 & 1.42 & 5.97 & 0.055 \\
    \textit{w/o multi-scenario data} & 6.26 & 1.99 & 18.77 & 4.81 & 0.76 & 17.81 & 4.85 & 1.04 & 4.62 & 0.040\\
    \midrule
    Ours & \textbf{5.83} & \textbf{1.89} & \textbf{18.41} & \textbf{4.67} & \textbf{0.71} & \textbf{17.12} & \textbf{4.75} & \textbf{0.98} & \textbf{4.30} & \textbf{0.038} \\
    \bottomrule[1.2pt]
    \end{tabular}
    \vspace{0.2cm}
\end{table*}

\subsection{Ablation study}
To better understand the contribution of each component in our framework, we conduct a series of ablation experiments.

\noindent\textbf{Single cross-attention.}
To validate the importance of utilizing two cross-attention layers, we replaces our dual cross-attention design with a single cross-attention layer that receives a mixed audio stream (speaker A + speaker B).
As shown in \cref{tab:ablation}, this leads to a substantial degradation in speaking performance, especially in lip-synchronization metrics such as LVE and MHD. This decline is due to the difficulty in differentiating between the speech features when processing the mixed audio stream.

\noindent\textbf{Without stage 1.}
This variant removes the audio-free training stage and directly trains the model using Stage 2 only.
Without stage 1, the model struggles to generate natural listening behaviors, resulting in a clear drop in listening performance across FDD, PDD, JDD, and FID.
These results confirm that learning the internal motion prior plays a vital role for natural listening reactions.

\noindent\textbf{Without multi-scenario data.}
In this variant, we train the audio-free generator in stage 1 using only the conversational data from stage 2, removing the diverse multi-scenario dataset.
As shown in \cref{tab:ablation}, this leads to consistent drops across both speaking and listening metrics.
Speaking performance becomes less stable, with higher LVE and MHD.
Listening performance also declines noticeably, reflected by increased dynamic deviations and worse FID.
These results indicate that multi-scenario data enables the model to learn a stronger internal motion prior, which generalizes better to both expressive speaking and listening behaviors.

\begin{table}
  \caption{Comparison of real-time performance and computational costs. ``RT'' indicates whether real-time use is supported.}
  \label{tab:real-time}
  \centering
  \small
  \renewcommand\arraystretch{1.2}
  \tablestyle{8.5pt}{1.05}
  \begin{tabular}{lcccc}
    \toprule[1.2pt]
    Method & RT & FPS & GFLOPS &  \#Params. \\
    \midrule
    ARTalk* \cite{artalk2025} & \correct & 357.7 & 720.7 & 489.5M\\
    DualTalk \cite{peng2025dualtalk} & \wrong & N/A & 348.1 & 647.3M\\
    \proposed (ours) & \correct & 560.6 & 739.4 & 421.3M \\
    \bottomrule[1.2pt]
  \end{tabular}
\end{table}

\subsection{Real-time performance}
\cref{tab:real-time} compares our method with ARTalk \cite{artalk2025} and DualTalk \cite{peng2025dualtalk} in terms of computational cost and real-time capability.
All experiments are conducted on a single NVIDIA GeForce RTX 5090 GPU.
Although \proposed requires higher GFLOPS, it achieves the highest frame rate at 560.6 FPS, significantly outperforming both ARTalk and DualTalk (N/A due to its non–end-to-end design).
Overall, \proposed provides the best trade-off: real-time generation, high throughput, and a smaller model size, making it well-suited for practical applications.

\section{Conclusion and discussion}
In this paper, we present \proposed, the first end-to-end audio-driven framework for generating unified speak–listen facial motions. Through a detailed analysis of audio–motion correlation, we identify the fundamental cause of listening stiffness in conventional end-to-end training. Building on this insight, we introduce a two-stage training paradigm that separates the learning of internal motion priors from audio-driven modulation.
Extensive experiments on large-scale conversational datasets demonstrate that \proposed achieves state-of-the-art speaking accuracy, the most diverse and natural listening expressions, and real-time performance.
Overall, \proposed establishes a foundation for conversational digital humans and opens new possibilities for interactive, lifelike human–AI communication.

\noindent\textbf{Limitations and future work.}
While \proposed demonstrates strong capability in jointly generating speaking and listening motions, it has several limitations.
First, the model lacks semantic understanding, so listener behaviors are primarily driven by acoustic cues (\eg, rhythm, emphasis) rather than conversational meaning.
As a result, it cannot yet produce semantically grounded reactions, such as nodding to agree or shaking the head to disagree.
In addition, our chunk-based formulation limits fully continuous motion generation, occasionally causing subtle discontinuities.
We view these challenges as promising directions for future work.
Incorporating semantic information and developing continuous long-horizon models may enable more context-aware and smoother conversational avatars.

%% file: sec/X_suppl.tex
\clearpage
\setcounter{page}{1}
\maketitlesupplementary

\begin{table*}[b!]
    \caption{
        Quantitative evaluation of the audio-free generator trained in stage 1. 
    }
    \label{tab:free-gen}
    \centering
    \small
    \renewcommand\arraystretch{1.1}
    \begin{tabular}{l|ccccc|ccccc}
    \toprule[1.2pt]
    & \multicolumn{5}{c|}{Speak} & \multicolumn{5}{c}{Listen} \\
    Method      & LVE$\downarrow$ & MHD$\downarrow$  & FDD$\downarrow$  & PDD$\downarrow$ & JDD$\downarrow$ & FDD$\downarrow$  & PDD$\downarrow$ & JDD$\downarrow$ & F-FID$\downarrow$  & P-FID$\downarrow$  \\
    \midrule
    Audio-free generator (stage 1 only) & - & - & 23.97 & 5.61 & 1.34 & 21.83 & 5.17 & 1.59 & 6.45 & 0.053\\
    \proposed (stage 1 + stage 2) & 5.83 & 1.89 & 18.41 & 4.67 & 0.71 & 17.12 & 4.75 & 0.98 & 4.30 & 0.038 \\
    \bottomrule[1.2pt]
    \end{tabular}
    \vspace{0.3cm}
\end{table*}

\begin{figure*}[b!]
\begin{center}
\centerline{\includegraphics[width=0.95\linewidth]{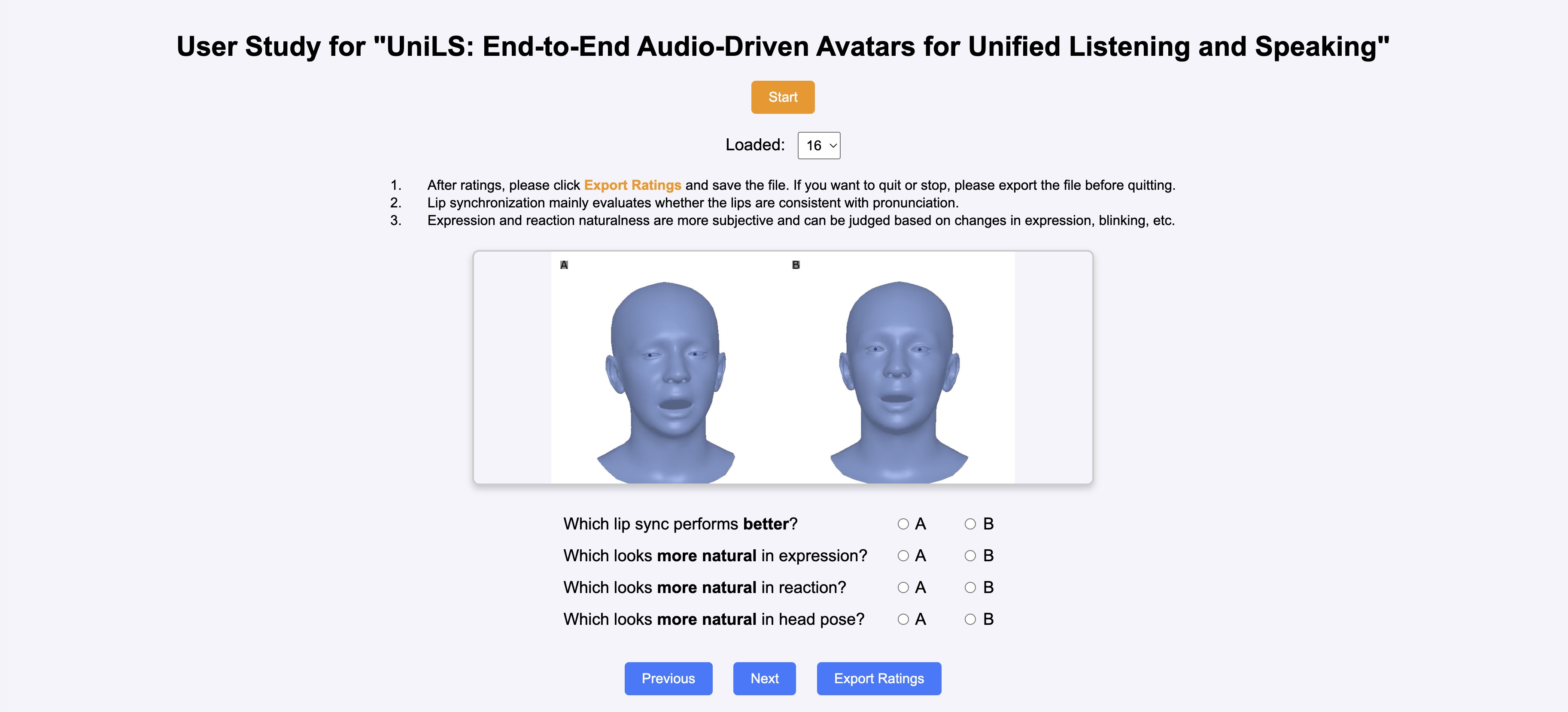}}
\caption{
The interface of our user study. Users evaluate each video based on four perspectives: lip synchronization, expression naturalness, reaction naturalness, and head pose naturalness. All these four perspectives are judged by comparing methods A and B. One of the videos (A or B) is generated by our method, and the other by a baseline method, with their order randomized.
}
\label{fig:us_interface}
\end{center}
\end{figure*}

\section{Evaluation of audio-free generator}
We evaluate the motion produced by the audio-free generator trained in stage 1. 
During evaluation, the stage-1 model is tested in a free autoregressive manner, where only the initial motion chunk is provided and all subsequent motions are generated without any audio guidance.
As shown in \cref{tab:free-gen}, we compare its outputs with the full two-stage \proposed model across both speaking and listening metrics.
Although the audio-free generator receives no audio input, it still produces motions with noticeable diversity and natural temporal variation, particularly in metrics related to dynamic deviation (FDD, PDD, JDD).
This shows that stage 1  effectively captures human’s internal motion priors, which prevents the model from collapsing into trivial or frozen outputs and instead encourages natural variability.

\section{User study details}
\cref{fig:us_interface} shows the interface of our user study. We collect responses from 25 participants, each completing 32 pairwise comparison trials (128 questions in total). All comparisons are performed on videos from the Seamless Interaction dataset. To avoid positional bias, the assignment of our method and the baseline to “A” or “B” is randomized for every trial.
Each comparison includes four single-choice questions. Following prior works \cite{faceformer2022,codetalker2023}, the first two assess lip synchronization and expression naturalness. We further introduce two questions targeting conversational behavior: reaction naturalness and head pose naturalness. For each question, participants select which video (A or B) appears more natural or better aligned with the audio.

\section{Ethics statement}
\proposed is designed to generate audio-driven 3D conversational avatars. While such technology has clear benefits in domains like education and human–computer interaction, it also presents potential ethical risks if misused, such as impersonation, generating misleading content, or producing avatars of real individuals without consent. We are aware of these concerns and strongly discourage any form of misuse. To mitigate these risks, we adopt the following safeguards:

\begin{itemize}
    \item \textbf{Visible watermarking.} We add a clear watermark to all synthesized videos so that viewers can immediately identify them as generated content, reducing the risk of deception.
    \item \textbf{Restricted identity usage.} We limit the target identities to virtual characters (\eg, virtual idols) and prohibit synthesizing real individuals without formal consent. Any generated content may only be used for educational or other legitimate purposes, and misuse is subject to accountability as described below.
    \item \textbf{Invisible watermarking and provenance.} We inject invisible watermarks into synthesized videos to record the source (\eg, the producer’s IP), ensuring traceability and encouraging creators to carefully consider ethical implications before generating content.
\end{itemize}

\noindent In summary, we provide strict licensing terms and technical measures to minimize the risk of abuse of \proposed. Broader efforts from researchers, practitioners, policymakers, and users are needed to ensure responsible development and use of avatar-generation technologies. With appropriate safeguards and ethical use, \proposed offers meaningful benefits for a wide range of real-world applications.